\documentclass[10pt,twocolumn,letterpaper]{article}

\usepackage{iccv}
\usepackage{times}
\usepackage{epsfig}
\usepackage{graphicx}
\usepackage{amsmath}
\usepackage{amssymb}

\usepackage[x11names]{xcolor}
\usepackage{enumitem}
\usepackage{float}
\usepackage{multicol}
\usepackage{booktabs}
\usepackage{multirow}
\usepackage{colortbl}
\usepackage{tabu}
\iccvfinalcopy

\newcommand{\red}[1]{\textcolor{red}{#1}}

\usepackage[pagebackref=true,breaklinks=true,letterpaper=true,colorlinks,bookmarks=false]{hyperref}

\usepackage{cleveref}

\def\iccvPaperID{6159} 

\ificcvfinal\fi

\usepackage{lipsum}

\newcommand\blfootnote[1]{%
  \begingroup
  \renewcommand\thefootnote{}\footnote{#1}%
  \addtocounter{footnote}{-1}%
  \endgroup
}

\begin{document}

\title{Harvest Video Foundation Models via Efficient Post-Pretraining}

\author{
Yizhuo Li$^{1,2*}$, Kunchang Li$^{2,3*}$, Yinan He$^2$, Yi Wang$^2$, \\
Yali Wang$^{2,3}$, Limin Wang$^{2,4}$, Yu Qiao$^{2,3}$, Ping Luo$^{1,2}$ \\
\small $^1$The University of Hong Kong, $^2$Shanghai AI Lab, 
\small $^3$Shenzhen Institutes of Advanced Technology, CAS \\
\small $^4$State Key Laboratory for Novel Software Technology, Nanjing University
}


\maketitle
\blfootnote{This paper is partially supported by the National Key R\&D Program of China No.2022ZD0161000.}
\ificcvfinal\thispagestyle{empty}\fi

\newcommand{\todo}{\textcolor{red}{TODO\ }}
\begin{abstract}
    Building video-language foundation models is costly and difficult due to the redundant nature of video data and the lack of high-quality video-language datasets.
    In this paper, we propose an efficient framework to harvest video foundation models from image ones.  
    Our method is intuitively simple by randomly dropping input video patches and masking out input text during the post-pretraining procedure.
    The patch dropping boosts the training efficiency significantly and text masking enforces the learning of cross-modal fusion.
    We conduct extensive experiments to validate the effectiveness of our method on a wide range of video-language downstream tasks including various zero-shot tasks, video question answering, and video-text retrieval.
    Despite its simplicity, our method achieves state-of-the-art performances, which are comparable to some heavily pretrained video foundation models.
    Our method is extremely efficient and can be trained in less than one day on 8 GPUs, requiring only WebVid-10M~\cite{frozen} as pretraining data.
    We hope our method can serve as a simple yet strong counterpart for prevalent video foundation models, provide useful insights when building them, and make large pretrained models more accessible and sustainable.
    This is part of the InternVideo project \url{https://github.com/OpenGVLab/InternVideo}.

\end{abstract}

\section{Introduction}
\label{sec:intro}

A new line of research is rising in multimodal modeling which connects images and text with the help of cross-modal contrastive learning~\cite{radford2021learning, li2021align, jia2021scaling, yu2022coca}.
\blfootnote{*Interns at Shanghai AI Laboratory}
Large image foundation models like CLIP~\cite{radford2021learning} are capable of aligning images and text into one shared embedding space, demonstrating a powerful ability to model both visuals and languages. 
They deliver excellent performance on downstream tasks, especially zero-shot ones without further fine-tuning. 
Following the success of cross-modal modeling in the image-language domain, video foundation models have also flourished through training on massive video-text pairs.

\begin{figure}[ht]
    \centering
    \includegraphics[width=\linewidth]{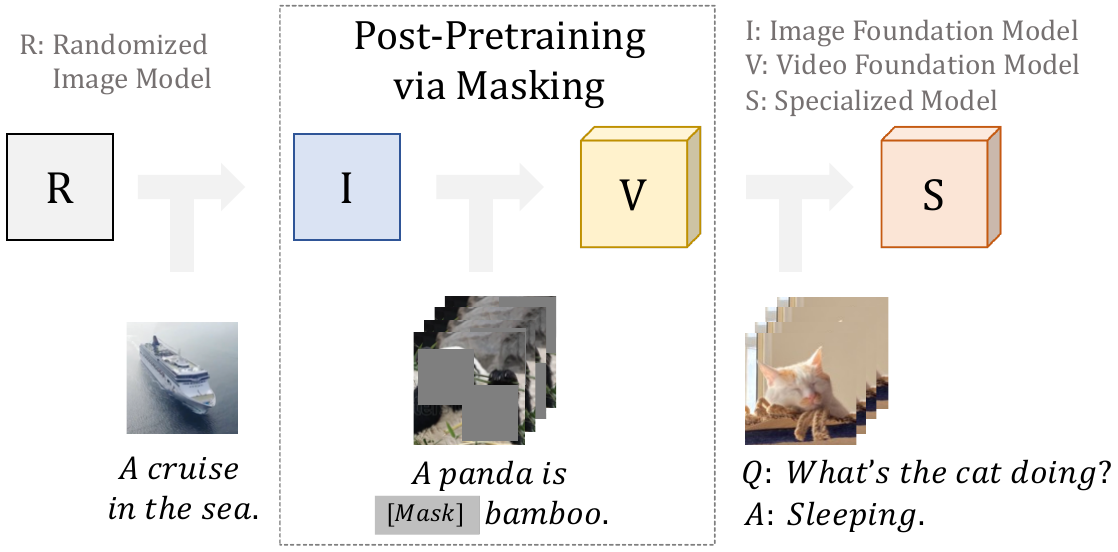}
    \caption{
    \textbf{The pipeline of our post-pretraining framework.}
    Our method explores the possibility of building powerful video foundation models upon image foundation models via post-pretraining.
    We employ a masking strategy on both video and text to boost the efficiency of post-pretraining and promote cross-modal fusion.
    }
    \label{teaser}
    \vspace{-0.3cm}
\end{figure}

However, building video foundation models can be costly and difficult. 
Video processing costs are directly proportional to the length of the video, which makes it more expensive than processing images. 
Successive frames in videos also tend to contain a lot of redundant spatial information, which wastes a lot of computation. 
Additionally, the existing video-text datasets, such as WebVid-10M~\cite{frozen}, are relatively small compared to their image-text counterparts (e.g., LAION-2B~\cite{schuhmann2022laion}), 
which makes constructing video foundation models more challenging.
In such scenarios, it is often more feasible and cost-effective to develop video foundation models based on existing image foundation models. 
Attempts to achieve this have been explored in BridgeFormer~\cite{ge2022bridging}, CLIP4Clip~\cite{clip4clip}, and CLIP-ViP~\cite{xue2022clip}. 
In this work, we aim to further push the limits of post-pretraining in an efficient manner.


We propose a simple post-pretraining framework to build video foundation models upon image foundation models. 
Our method follows the popular MAE paradigm~\cite{he2022masked,tong2022videomae,han2022turbo,li2022scaling}, by randomly dropping the input video patches with a certain probability.
We call it ``dropping'' instead of ``masking'' because we do not recover dropped patches or replace them with special tokens. However, we may use these terms interchangeably for convenience.
Additionally, we also randomly mask the input text and predict the masked token with an extra decoder. 
Video patch dropping and text masking look alike yet are applied for different purposes. Video patch dropping is designed to significantly boost training efficiency, while text masking is designed to promote modalities fusion to build a more capable video foundation model.
The framework is illustrated in Figure~\ref{teaser}.

We employ a straightforward post-pretraining procedure, by jointly optimizing contrastive loss and masked text prediction loss, trained on WebVid-10M~\cite{frozen} for only 50k steps.
We conduct extensive experiments on a wide range of video-language downstream tasks to evaluate the performance of our given framework, including multiple zero-shot tasks, video question answering, and video-text retrieval. 
Despite the simplicity of our method, we achieve SOTA performance comparable to popular video foundation models.
Moreover, our method is highly efficient and the post-pretraining procedure takes less than 192 GPU hours (using A100). 
As a comparison, a typical video foundation model like All-in-one~\cite{all_in_one} requires more than 5k GPU hours (using A100) with inferior performance.

Based on the experimental results, we give an in-depth discussion on existing paradigms for video foundation models. 
We attribute the effectiveness of our method to the powerful CLIP pretraining, which reveals that image-trained models can perform well on video-language tasks with an inexpensive post-pretraining procedure.
This reveals the limitation of existing video-language datasets, which may not provide enough temporal textual description to model the rich information in videos.
We also find that the text encoder plays a vital role in video-language tasks.
However, current video-language datasets may not be diverse or of high enough quality to train an adequate text encoder.
We hope our method can serve as a strong yet efficient counterpart for video foundation models and provides useful insights into building them.

\begin{figure*}[ht]
    \centering
    \includegraphics[width=\linewidth]{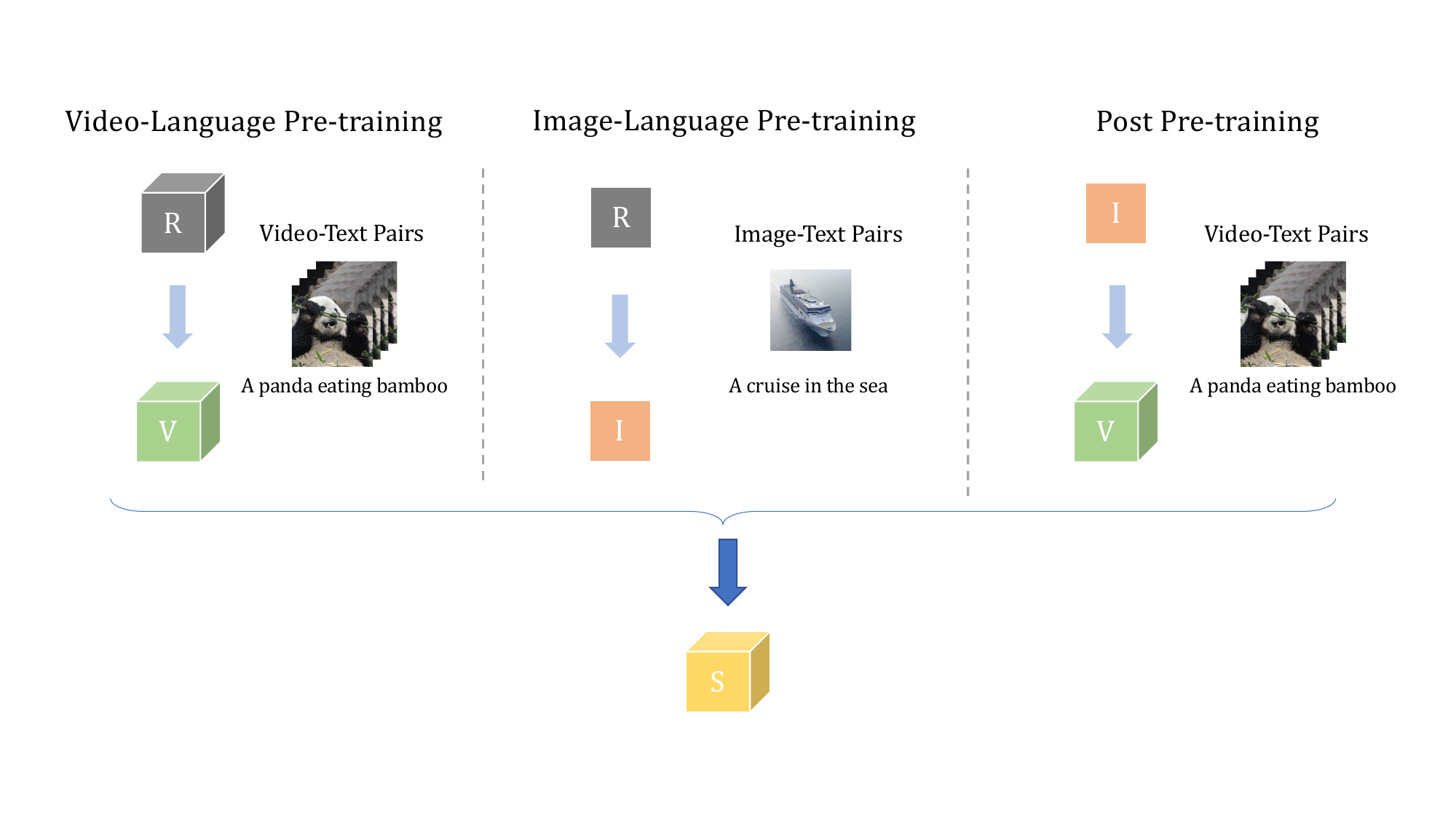}
    \caption{
        \textbf{Overall framework of the proposed method.} Our method is intuitively simple with video patch dropping and text masking design. 
        \textbf{1)} We randomly drop input video patches with a certain probability before feeding them into the visual encoder, without recovering them.
        \textbf{2)} We randomly replace a certain portion of input text tokens with special \texttt{[MASK]} token, and predict the masked targets by introducing a text decoder.
        \textbf{3)} Our method is built upon pre-trained CLIP model and keeps the text encoder frozen. The framework is jointly trained with the contrastive loss $\mathcal{L}_{con}$ and the masked language loss $\mathcal{L}_{mask}$.
    }
    \vspace{-0.3cm}
    \label{fig:framework}
\end{figure*}

\section{Related Work}
\label{sec:related}

\noindent 
\textbf{Video-Language Pretraining.}
Starting from the rapid development of image-language pretraining~\cite{LXMERT,uniter}, 
large-scale video-language pretraining with fine-tuning on specific downstream tasks has become the standard paradigm in the video-language understanding~\cite{merlot,violet,videoclip}. 
The earliest methods~\cite{videobert,actbert} directly extract the offline video and text features from well-pretrained visual and language encoders,  
while the recent methods~\cite{clipbert,frozen,reserve,all_in_one,singularity} have demonstrated the feasibility of end-to-end training.  
Besides, 
the popular methods often include two or three pretraining tasks,
e.g., masked language modeling~\cite{vl_bert,lavender},  
frame order modeling~\cite{merlot},
video-text matching~\cite{clipbert}, 
video-text contrastive learning~\cite{videoclip} and video-text masked modeling~\cite{violet}.
As for the training data,
the previous methods mainly leverage image-text pairs,
such as COCO Caption~\cite{coco}, Google Conceptual Captions~\cite{CC3M}, and Visual genome~\cite{vg}.
For better video-language understanding,
large-scale video-text pairs are introduced,
including WebVid-2M\cite{frozen}, HowTo100M~\cite{howto100m}, and YT-Temporal-180M~\cite{merlot}. 
Unlike most methods that require large-scale datasets or enormous training resources, our method requires only WebVid-10M~\cite{frozen} and 8 GPUs and can be trained in less than one day.

\noindent 
\textbf{Video-Language Downstream Tasks.}
Video-Language understanding tasks~\cite{value,liu2019use,jiang2020divide,le2020hierarchical,patrick2020support,gabeur2020multi} have attracted rapidly growing attention in the computer vision community and natural language processing community. 
In the period before the video-language pretraining booming, some specific downstream tasks have been widely studied including video question answering~\cite{tgif_qa,tvqa,tvqa+,xu2017video}, video-to-text retrieval~\cite{msrvtt,densecaption,tvr}, video caption~\cite{msvd,msrvtt,vatex,movie_description,zhou2018towards} and temporal localization~\cite{didemo,Tall,densecaption,tvr}. 
In the research paradigm of these tasks, offline video feature extraction~\cite{xie2018rethinking,gao2018motion,lei2021detecting,fan2019heterogeneous} plays an important role in performance.
With the rapid progress of video-language pretraining tasks~\cite{videobert,actbert,hero}, the performance of downstream tasks is further improved.
Recently, CLIP~\cite{radford2021learning} demonstrates very impressive transfer and generalization capacities in the video-language field,
including
video-text retrieval~\cite{clip4clip,clip2tv,clip2video,camoe},
video caption~\cite{clip4caption,clip_meet_video},
video summarization~\cite{clipit} and  zero-shot and few-shot recognition~\cite{coop,cocoop,tip-adapter}.
In this work,
we design a task adaption module to further extend CLIP for more diverse downstream tasks such as video question answering.

\noindent
\textbf{Masked Modeling.}
The prevailing masking modeling strategy in computer vision is introduced by MAE~\cite{he2022masked}. 
MAE randomly drops input vision tokens and reconstructs them as a proxy task to learn spatial representation. 
MAE-based methods have already been extended to videos~\cite{tong2022videomae,feichtenhofer2022masked,han2022turbo}. 
Our method follows a similar design but does not recover the masked tokens. 
Masked learning in natural language processing has a longer history back such as the highly influential work BERT~\cite{devlin2018bert}.
A recent work FLIP~\cite{li2022scaling} shares a similar idea with our method by applying masking to efficiently train cross-modal models but differs in several aspects. 
We discuss the differences in the following section.

\section{Methodology}
\label{sec:method}

Our method is a simple \textit{align before fuse} framework, which applies masking to both video and text inputs.
The method consists of three main components: (1) a video encoder, (2) a text encoder, and (3) a cross-modal fusion module.
As the mission of our method is to unleash the potential of image foundation models on video-language tasks via post-pretraining, we follow the paradigm of ALBEF~\cite{li2021align} and CoCa~\cite{yu2022coca} to append a modality-fusion module after video-language alignment.
By following a common paradigm similar to prevalent video foundation models, we demonstrate that our method can achieve superior performance without the need for specialized designs. The framework is illustrated in Figure~\ref{fig:framework}. 

\textbf{Video Patch Dropping.}
We apply video patch dropping following MAE-based methods~\cite{he2022masked,tong2022videomae,han2022turbo,li2022scaling}.
Videos in nature are temporally redundant~\cite{tong2022videomae} and thus require a lot more unnecessary computational resources to process.
During training, cross-modal contrastive learning also requires a relatively larger batch size compared to typical supervised video tasks for better performance~\cite{radford2021learning,li2022scaling}.
Therefore, we introduce video patch dropping to reduce the computational cost and meet the batch size requirement. 
Video patch dropping is a key component of our method as it alleviates the computational cost to a large extent.
We call it ``dropping'' instead of ``masking'' because we do not recover the dropped patches as in MAE~\cite{he2022masked} or VideoMAE~\cite{tong2022videomae}.
And we call it ``patch'' instead of ``token'' to distinguish it from the following text token masking.

\textbf{Text Masking.} We also apply masking on input text to create a proxy task for cross-modal fusion. 
In contrast to FLIP~\cite{li2022scaling}, which applies the same approach of dropping text tokens as video patches, our method employs a random replacement technique like BERT~\cite{devlin2018bert}, whereby a subset of text tokens is substituted with a designated token referred to as \texttt{[MASK]}. 
We employ a cross-modal transformer decoder as the multi-modal fuser.
The fuser is a standard transformer decoder. At cross-attention layers, the decoder takes text features as queries and video features as keys and values.
Given masked video features and text features, the target of the decoder is to predict the masked text tokens.
The decoder shares the same objective as the multimodal encoder in ALBEF~\cite{li2021align} or the captioner in CoCa~\cite{yu2022coca}, to push the ability of contrastive models beyond alignment.
With text masking as an auxiliary task, we can enforce the model to learn more fine-grained cross-modal information instead of only global semantics.
In this way, the model will perform better at tasks requiring modalities fusion such as video question answering without large-scale re-training.

\textbf{Training Objectives}
Our framework is optimized toward two objectives: video-text alignment and masked language modeling.
Video-text alignment is trained by minimizing the InfoNCE loss with global video and text features.
Masked language modeling is trained by minimizing the cross-entropy loss between the decoder's output and the ground-truth masked text tokens.

Our method shares similar ideas with a recent concurrent work FLIP~\cite{li2022scaling} but differs in the following aspects: \textbf{(i)} Our method is aimed at harvesting the potential of pretrained image foundation models on video-language tasks via post-pretraining, while FLIP is aimed at speeding up CLIP training. Videos in nature are more redundant than images and benefit more from patch dropping. \textbf{(ii)} Our method employs a cross-modal transformer decoder as the multi-modal fuser, while FLIP does not involve any cross-modal fusion. The decoder pushes the ability of contrastive models beyond alignment and makes our method generalize on more downstream tasks. \textbf{(iii)} FLIP requires an unmasking procedure before being applied to downstream tasks while our method does not, which makes our method more efficient.
\section{Experiments}
\label{sec:exp}

In this section, we first describe the settings for post-pretraining in Sec.~\ref{sec:ppt} and then demonstrate the performance of the post-pretrained model by evaluating on a variety of downstream tasks.
In Sec.~\ref{sec:zero-shot}, we validate the effectiveness of our method on zero-shot tasks. We evaluate by fine-tuning on video-text retrieval in Sec.~\ref{sec:vt} and video question answering in Sec.~\ref{sec:vqa}. Ablations and discussions are conducted in Sec.~\ref{sec:ablation} and Sec.~\ref{sec:discussion}.

\subsection{Post-Pretraining}
\label{sec:ppt}

\noindent\textbf{Dataset.} 
We choose WebVid-10M~\cite{frozen}, a diverse and clean video-text dataset collected from stock footage sites consisting of 10.7M video-text pairs. 
Compared with typical video foundation models, it is $1/10$ of HD-VILA-100M~\cite{xue2022advancing} used in CLIP-VIP~\cite{xue2022clip}, $1/10$ of WebVid-2.5M + Howto100M used in All-in-One~\cite{all_in_one}, and $1/18$ of YT-Temporal-180M used in MERLOT~\cite{merlot}.
No additional data or pretrained models are used other than WebVid-10M and CLIP.

\noindent\textbf{Architecture.} We use a simplified UniformerV2~\cite{anonymous2023uniformerv} as the visual encoder by default.
Since the spatiotemporal convolution in UniformerV2 hinders the utilization of video patch dropping,
we only insert the global UniBlocks but remove the Dynamic Position Encoding module.
We initialize additional parameters in a way that the output is identical to the original CLIP model which we find to be essential for decent zero-shot performance. 
The masked language module is a standard 4-layer transformer~\cite{transformer} decoder with a dimension of 512 followed by a two-layer MLP. 
Other settings leave CLIP Base/16 untouched. 
We ablate the choice of the visual encoder by comparing with the vanilla ViT\cite{Dosovitskiy2021AnII} backbone. 
Unlike UniformerV2 with the temporal modules, the features of ViT are extracted frame-wise and directly averaged across frames.
UniformerV2 endows a stronger temporal modeling ability with extra modules. Comparing UniformerV2 with ViT will reveal the impacts of explicit temporal modeling on different datasets and tasks.

\noindent\textbf{Training.} Thanks to the efficiency of patch dropping, we can post-pretrain with minimal computational resources. By default, we train for 50k steps on 8 A100 GPUs within 1 day. As a comparison, a typical video foundation model like All-in-one~\cite{all_in_one} requires 32 A100 GPUs for 7 days. 
The model is trained with a batch size of 1024, a learning rate of $1\times 10^{-5}$, weight decay of 0.2, and a cosine annealing schedule with 4k warm-up steps. 
The text encoder is frozen during post-pretraining as the original training corpus of CLIP is much richer than WebVid-10M~\cite{frozen}. CLIP-ViP~\cite{xue2022clip} also demonstrates that there exists a domain gap between the pretraining data and downstream tasks. Without additional data or a pretrained captioner, freezing the text encoder is the optimal choice.


\noindent\textbf{Implementation.} We follow VideoMAE~\cite{tong2022videomae} using a large dropping ratio of $90\%$ for video input, saving computational resources to a large extent. We randomly sample 8 frames per clip as video input. For text input, we use a mask ratio of $15\%$ by default following BERT~\cite{devlin2018bert}. The effects of different dropping ratios and mask ratios are ablated in Sec.~\ref{sec:ablation}.

\subsection{Zero-Shot Tasks}
\label{sec:zero-shot}

We first validate the effectiveness of our method on zero-shot tasks. One of the key challenges of zero-shot learning is the distribution shift between the pretraining and target domains. CLIP-based methods suffer more from overfitting as the pretraining dataset is usually much larger than the post-pretraining dataset~\cite{xue2022clip}. WiSE-FT~\cite{wortsman2022robust} tackles this problem by simply weighted ensembling the pretrained model and the fine-tuned model. Unlike in WiSE-FT where the target domain is also an image one, our scenario requires a more flexible method to solve the domain gap. We extend WiSE-FT to an online and multiple-checkpoint version. In short, we evenly ensemble using $l$ checkpoints every $k$ epochs/intervals during post-pretraining. We provide a detailed explanation in the supplementary and validate this design in the ablation study. By default, we set $k=10$ and $l=3$. We do not apply patch dropping in zero-shot tasks as patch dropping hurts the performance of zero-shot tasks greatly without an unmasking procedure.

\begin{table}
    \centering
    \small
    \begin{tabular}[htop]{l c}
        \toprule
        Method  & Top-1 Accuracy \\
        \midrule
        ER-ZSAR~\cite{chen2021elaborative}  & 42.1 \\
        ActionCLIP~\cite{actionclip} & {56.0} \\
        \hline
        Ours without WiSE-FT & 54.0 \\
        Ours with ViT & \textbf{56.8} \\
        Ours & \underline{56.7} \\
        \bottomrule
    \end{tabular}
    \caption{\textbf{Zero-shot action recognition on Kinetics-400. } Despite our effort to tackle distribution shift by freezing text encoder, the performance is still lower than SOTA method without massive additional data or interfering with the weights.}
    \label{tab:zs_ar}
\end{table}

\begin{table}
    \centering
    \small
    \begin{tabular}[htop]{l c c}
        \toprule
        Method  & MSRVTT  & LSMDC \\
        \midrule
        \color{gray}{JSFusion}~\cite{yu2018joint}  & \color{gray}{83.4} & \color{gray}{73.5} \\
        \color{gray}{All-in-one}~\cite{all_in_one} & \color{gray}{92.3} & \color{gray}{84.4} \\
        \color{gray}{MERLOT}~\cite{merlot} & \color{gray}{90.9} & \color{gray}{81.7} \\
        \color{gray}{VIOLET}~\cite{violet} & \color{gray}{91.9} & \color{gray}{82.8} \\
        All-in-one~\cite{all_in_one} & 80.3 & 56.3\\
        \hline
        Ours without WiSE-FT & {92.6}  & {74.9} \\
        Ours with ViT & \textbf{93.5} & \textbf{76.5} \\
        Ours & \underline{93.2} & \underline{76.3} \\
        \bottomrule
    \end{tabular}
    \caption{\textbf{Zero-shot multiple-choice on MSRVTT and LSMDC.} 
    Those methods with supervised training are {\color{gray}{grayed out}}. 
    Unlike action recognition, our method surpasses the SOTA method even without WiSE-FT. This is attributed to the frozen text encoder as multiple-choice task requires better textual modeling.}
    \vspace{-0.3cm}
    \label{tab:zs_mc}
\end{table}

\begin{table}
    \centering 
    \small
    \setlength\tabcolsep{3pt}
    \begin{tabu}[htop]{l c c c c c c c c c c c c c c c c c c c c c}
        \toprule
        Method & R@1$\uparrow$ & R@5$\uparrow$ & R@10$\uparrow$ & MdR$\downarrow$ \\
        \midrule
        & \multicolumn{4}{c}{MSR-VTT} \\
        \rowfont{\color{gray}}VideoCLIP~\cite{videoclip}  & 10.4 & 22.2 & 30 & - \\
        \rowfont{\color{gray}}Frozen~\cite{frozen} & 24.7 & 46.9 & 57.2 & 7.0  \\
        \rowfont{\color{gray}}{BridgeFormer}~\cite{ge2022bridging} & 33.2 & 58.0 & 68.6 & 4.0 \\
        ALPRO~\cite{li2022align} & 24.1 & 44.7 & 55.4 & 8.0 \\
        VIOLET~\cite{violet} & 25.9 & 49.5 & 59.7 & -  \\
        OmniVL\red{$\dag$}~\cite{wang2022omnivl} & \underline{34.6} & \underline{58.4} & \underline{66.6} & - \\
        \hline
        Ours with ViT & 32.6 & 54.9 & 65.5 & {4.0} \\
        Ours & \textbf{36.2} & \textbf{60.3} & \textbf{69.7} & {3.0} \\
        \midrule
        & \multicolumn{4}{c}{MSVD} \\
        \rowfont{\color{gray}}NoiseEst~\cite{amrani2021noise} & 13.7 & 35.7 & 47.7 & 12.0 \\
        \rowfont{\color{gray}}Frozen~\cite{frozen} & 33.7 & 64.7 & 76.3 & 3.0 \\
        \rowfont{\color{gray}}BridgeFormer~\cite{ge2022bridging}  & {48.4} & {76.4} & {85.8} & {2.0} \\
        \hline
        Ours with ViT & \underline{39.2} & \underline{66.9} & \underline{76.2} & 2.0 \\
        Ours & \textbf{44.0} & \textbf{72.7} &  \textbf{82.5} & {2.0} \\
        \midrule
        & \multicolumn{4}{c}{LSMDC} \\
        \rowfont{\color{gray}}NoiseEst~\cite{amrani2021noise} & 4.2 & 11.6 & 17.1 & 119.0 \\
        \rowfont{\color{gray}}Frozen~\cite{frozen} & 9.3 & 22.0 & 30.1 & 51.0 \\
        \rowfont{\color{gray}}BridgeFormer~\cite{ge2022bridging} & 15.5 & {30.7} & {38.7} & {22.0} \\
        \hline
        Ours with ViT & \textbf{17.5} & \underline{29.9} & \textbf{38.0} & {25.0} \\
        Ours & \underline{16.0} & \textbf{30.2} &  \underline{36.7} & {28.0} \\
        \midrule
        & \multicolumn{4}{c}{DiDeMo} \\
        \rowfont{\color{gray}}VideoCLIP~\cite{videoclip} & 16.6 & 46.9 & - & - \\
        \rowfont{\color{gray}}Frozen~\cite{frozen} & 20.2 & 46.4 & 58.5 & 7.0  \\
        \rowfont{\color{gray}}BridgeFormer~\cite{ge2022bridging} & 25.6 & 50.6 & 61.1 & 5.0 \\
        VIOLET~\cite{violet} & 23.5 & 49.8 & 59.8 & - \\
        ALPRO~\cite{li2022align} & 23.8 & 47.3 & 57.9 & {3.0} \\
        OmniVL\red{$\dag$}~\cite{wang2022omnivl} & \textbf{33.3} & \textbf{58.7} & \textbf{68.5} & - \\
        \hline
        Ours with ViT & 29.8 & 54.3 & \underline{63.6} & {5.0} \\
        Ours & \underline{32.2} & \underline{58.0} & \textbf{68.5} & {4.0} \\
        \midrule
        & \multicolumn{4}{c}{VATEX} \\
        \rowfont{\color{gray}}CLIP~\cite{radford2021learning} & {39.7} & {72.3} & {82.2} & {2.0} \\
        Ours with ViT & \underline{45.2} & \underline{76.5} & \underline{85.4} & 2.0\\
        Ours & \textbf{48.9} & \textbf{80.6} & \textbf{88.4} & {2.0} \\
        \bottomrule
    \end{tabu}
    \caption{\textbf{Zero-shot video-text retrieval on MSR-VTT, MSVD, LSMDC, DiDeMo, and VATEX.}
    Video foundation models and retrieval-specialized methods are mixed for reference.
    Our method is aimed to compare with video foundation models for general purpose, therefore those methods specially designed for retrieval are {\color{gray}{grayed out}}.
    ``\red{$\dag$}'' utilizes matching loss to rerank the retrieved results for better performance.
    Results without WiSE-FT are not reported because they all fail in zero-shot video-text retrieval.
    }
    \vspace{-0.3cm}
    \label{tab:zs_vt}
\end{table}

\noindent\textbf{Zero-Shot Action Recognition.}
We report zero-shot action recognition performance on Kinetics-400\cite{Smaira2020ASN} as an indicator of zero-shot classification ability. We follow the setting of ActionCLIP~\cite{actionclip} with textual prompt and average the similarity between the normalized visual classification token and text classification tokens. The results are reported in Table~\ref{tab:zs_ar}. Despite our effort to tackle the distribution shift by freezing the text encoder, the performance after post-pretraining drops to 54.0\% top-1 accuracy. With the modified WiSE-FT, the performance is boosted to 56.0\% top-1 accuracy, indicating that the distribution shift is alleviated without massive additional data. Our method shows slightly better performance with ViT backbone, which indicates less need for temporal information in this scenario. 

\noindent\textbf{Zero-Shot Multiple-Choice.} 
We evaluate the performance on the zero-shot multiple-choice task. 
The objective of the multiple-choice task is to find the correct caption from the candidates, serving as a simplified version of the retrieval task. 
We report zero-shot performance on MSRVTT~\cite{yu2018joint}, and LSMDC~\cite{lsmdc_mc} in Table~\ref{tab:zs_mc}.
Our method even outperforms some supervised methods with such a simple framework.
Similar to action recognition, using ViT as the backbone yields better results due to less need for temporal information.
This task presents a smaller relative gap between post-pretraining with WiSE-FT and the one without.
We attribute this to the fact that the multiple-choice task requires better textual modeling ability, which is retained via freezing the text encoder.
A similar observation is presented in Co-Tokenization~\cite{piergiovanni2022video}, in which a pretrained T5 model achieves almost $100\%$ accuracy on the multiple-choice task in TGIF-QA~\cite{tgif_qa} even without video.

\noindent\textbf{Zero-Shot Video-Text Retrieval.} 
We evaluate zero-shot video-text retrieval performance on 5 popular video-text retrieval datasets including MSRVTT~\cite{msrvtt}, MSVD~\cite{msvd}, LSMDC~\cite{movie_description}, DiDeMo~\cite{didemo}, and VATEX~\cite{vatex}. A brief introduction of these datasets can be found in Sec.~\ref{sec:vt}. The results are reported in Table~\ref{tab:zs_vt}.
Our method demonstrates superior performance across all 5 datasets. 
The only inferior case is DiDeMo on which the performance is only slightly lower than OmniVL, which uses an extra matching loss to rank the retrieved results while our method is purely similarity-based.
The different characteristic of video-text retrieval task is that without WiSE-FT, results are only single-digit, unlike classification and multiple-choice tasks.
This reveals that the distribution shift shows a non-negligible effect that is too large to be alleviated by only freezing the text encoder.

Our method shows superior performance on zero-shot tasks with the modified WiSE-FT. This is surprising yet expected. First, CLIP models are already powerful zero-shot models trained on diverse data, while those video-language tasks are still limited to a small domain. Second, we handle the distribution shift to a large extent by freezing the text encoder and using WiSE-FT. However, the failure on retrieval without WiSE-FT indicates that classification and multi-choice rely more on static vision, while retrieval requires more dynamic and interactive modeling, thus suffering more from the distribution shift. On several tasks, using the vanilla ViT as the backbone shows better performance. This may be due to that using static vision is sufficient to handle those benchmarks. We discuss this in Sec.~\ref{sec:discussion}.

\begin{table}
    \centering
    \small
    \setlength\tabcolsep{3pt}
    \begin{tabu}[ht]{l l l l l}
        \toprule
        Method & R@1$\uparrow$ & R@5$\uparrow$ & R@10$\uparrow$ & MdR$\downarrow$ \\
        \midrule
        & \multicolumn{4}{c}{MSR-VTT} \\
        \rowfont{\color{gray}}FROZEN~\cite{frozen} & 31.0 & 59.5 & 70.5 & 3.0 \\
        \rowfont{\color{gray}}CLIP4Clip~\cite{clip4clip} & 44.5 & 71.4 & 81.6 & 2.0 \\
        \rowfont{\color{gray}}BridgeFormer~\cite{ge2022bridging} & 44.9 & 71.9 & 80.3 & 2.0 \\
        \rowfont{\color{gray}}CLIP-ViP\red{$\dag$}~\cite{xue2022clip} & 54.2 & 77.2 & 84.8 & 1.0 \\
        ClipBERT~\cite{clipbert} & 22.0 & 46.8 & 59.9 & 6.0 \\
        ALPRO~\cite{li2022align} & 33.9 & 60.7 & 73.2 & 3.0 \\
        All-in-one~\cite{all_in_one} & 37.9 & 68.1 & 77.1 & - \\
        VIOLETv2\red{$\dag$}~\cite{fu2022empirical} & 37.2 & 64.8 & 75.8 & - \\
        OmniVL\red{$\dag$}~\cite{wang2022omnivl} & \textbf{47.8} & \textbf{74.2} & \textbf{83.8} & - \\
        \arrayrulecolor{gray}\hline\arrayrulecolor{black}
        Ours w/o dropping & 45.3 & 72.5 & 80.9 & 2.0 \\
        Ours with ViT & 45.7 & 73.8 & 82.1 & 2.0 \\
        Ours & \underline{47.4} & \underline{73.2} & \underline{82.6} & 2.0 \\
        \midrule
        & \multicolumn{4}{c}{MSVD} \\ 
        \rowfont{\color{gray}}FROZEN~\cite{frozen} & 33.7 & 64.7 & 76.3 & 3.0 \\
        \rowfont{\color{gray}}CLIP4Clip~\cite{clip4clip} & 46.2 & 76.1 & 84.6 & 2.0 \\
        \rowfont{\color{gray}}BridgeFormer~\cite{ge2022bridging} & 54.4 & 82.8 & 89.4 & 1.0 \\
        \arrayrulecolor{gray}\hline\arrayrulecolor{black}
        Ours w/o dropping & {49.9} & \underline{79.7} & {87.8} & 2.0 \\
        Ours with ViT & \underline{50.4} & 79.4 & \underline{88.0} & 1.0 \\
        Ours & \textbf{51.0} & \textbf{80.5} & \textbf{88.4} & {1.0} \\
        \midrule
        & \multicolumn{4}{c}{LSMDC} \\
        \rowfont{\color{gray}}FROZEN~\cite{frozen} & 9.3 & 22.0 & 30.1 & 51.0 \\
        \rowfont{\color{gray}}BridgeFormer~\cite{ge2022bridging} & 21.8 & 41.1 & 50.6 & 10.0 \\
        \rowfont{\color{gray}}CLIP4Clip~\cite{clip4clip} & 22.6 & 41.0 & 49.1 & 11.0 \\
        \rowfont{\color{gray}}CLIP-ViP\red{$\dag$}~\cite{xue2022clip} & 29.4 & 50.6 & 59.0 & \\
        VIOLETv2\red{$\dag$}~\cite{fu2022empirical} & \underline{24.0} & 43.5 & \underline{54.1} & - \\
        \arrayrulecolor{gray}\hline\arrayrulecolor{black}
        Ours w/o dropping & 22.5 & \underline{43.7} & \textbf{54.3} & 8.0 \\
        Ours with ViT & 22.0 & 42.3 & 52.5 & 9.0 \\
        Ours & \textbf{24.7} & \textbf{44.0} & {53.1} & 8.0 \\
        \midrule
        & \multicolumn{4}{c}{DiDeMo} \\ 
        \rowfont{\color{gray}}FROZEN~\cite{frozen} & 31.0 & 59.8 & 72.4 & 3.0 \\
        \rowfont{\color{gray}}BridgeFormer~\cite{ge2022bridging} & 37.0 & 62.2 & 73.9 & 3.0 \\
        \rowfont{\color{gray}}CLIP4Clip~\cite{clip4clip} & 43.4 & 70.2 & 80.6 & 2.0 \\
        \rowfont{\color{gray}}CLIP-ViP\red{$\dag$}~\cite{xue2022clip} & 50.5 & 78.4 & 87.1 & 1.0 \\
        ClipBERT~\cite{clipbert} & 21.1 & 47.3 & 61.1 & 6.3 \\
        All-in-one~\cite{all_in_one} & 31.2 & 60.5 & 72.1 & 3.0 \\
        ALPRO~\cite{li2022align} & 35.9 & 67.5 & 78.8 & 3.0 \\
        VIOLETv2\red{$\dag$}~\cite{fu2022empirical} & \underline{47.9} & \underline{76.5} & \underline{84.1} & - \\
        OmniVL\red{$\dag$}~\cite{wang2022omnivl} & \textbf{52.4} & \textbf{79.5} & \textbf{85.4}  & - \\
        \arrayrulecolor{gray}\hline\arrayrulecolor{black}
        Ours w/o dropping & {46.5} & {73.8} & {81.9} & 2.0 \\
        Ours with ViT & 45.4 & 72.4 & 80.4 & 2.0 \\
        Ours & {46.7} & {74.4} & {82.4} & 2.0 \\
        \midrule
        & \multicolumn{4}{c}{VATEX} \\
        \rowfont{\color{gray}}CLIP4Clip~\cite{clip4clip} & 55.9 & 89.2 & 95.0 & 1.0 \\
        \arrayrulecolor{gray}\hline\arrayrulecolor{black}
        Ours w/o dropping & \underline{64.4} & \textbf{92.2} & \underline{96.3} & 1.0 \\
        Ours with ViT & 64.2 & \underline{92.1} & \underline{96.3} & 1.0 \\
        Ours & \textbf{64.5} & \underline{92.1} & \textbf{96.5} & 1.0 \\
        \bottomrule
    \end{tabu}
    \caption{
    \textbf{Video-text retrieval task on MSR-VTT, MSVD, LSMDC, DiDeMo, and VATEX. }
    Our baseline model is CLIP4Clip~\cite{clip4clip} and our method only provides the pre-trained model. 
    Methods specially designed for retrieval are {\color{gray}{grayed out}}. 
    ``\red{$\dag$}'' marks those utilizing matching loss to rerank the retrieved results for better performance. 
    CLIP-ViP utilizes substantially more data but is highly related as a post-pretraining counterpart.
    }
    \label{tab:retrieval}
\end{table}

\subsection{Video-Text Retrieval}
\label{sec:vt}

The introduction of video patch dropping should naturally lead to two benefits: saving computational resources greatly and improving the performance of contrastive learning by fitting larger batch size~\cite{radford2021learning,li2022scaling}. We demonstrate the effectiveness of video patch dropping with the retrieval task which benefits more from large batch size.

\noindent\textbf{Datasets.}
We evaluate the performance of our method on 5 datasets including MSRVTT~\cite{msrvtt}, MSVD~\cite{msvd}, LSMDC~\cite{movie_description}, DiDeMo~\cite{didemo}, and VATEX~\cite{vatex}. MSRVTT contains 10,000 videos in total and 200,000 captions. MSVD contains 1,970 videos in total and 40 captions for each video. LSMDC contains 118,081 videos in total and each video has one caption. DiDeMo contains 10,000 videos in total and 40,000 captions. 

\noindent\textbf{Implementation and Training.}
We follow the standard data split in CLIP4Clip~\cite{clip4clip} and also follow its setting for fine-tuning on video-text retrieval tasks. 
When training without video patch dropping, we use a batch size of 24 per GPU due to memory limitation. The batch size is increased to 128 with video patch dropping as the memory requirement is significantly lifted.

The results are reported in Table~\ref{tab:retrieval} with R@1, R@5, R@10, and median rank.
It should be noted that the baseline of our method is CLIP4Clip and our method only provides the post-pretrained model. Some SOTA methods like OmniVL~\cite{wang2022omnivl}, VIOLETv2~\cite{fu2022empirical}, and CLIP-ViP~\cite{xue2022clip} utilize a matching loss when pretraining to rerank the retrieved results for better performance. While our method is purely similarity-based for generality and fair comparison. CLIP-ViP~\cite{xue2022clip} is a highly related work as a post-pretraining counterpart but uses substantially more data including 114.5M pairs and an additional pretrained captioner. 

With video patch dropping, our model achieves results comparable to SOTA methods on all datasets. We find that our method does not require an unmasking procedure to be applied on downstream tasks unlike FLIP~\cite{li2022scaling}. This is possibly due to the fact that our method is initialized with a pretrained CLIP and tries to alleviate distribution shift with a frozen text encoder and WiSE-FT. The performance gap between using a vanilla ViT and UniformerV2 as the backbone is smaller than the zero-shot setting. This can be attributed to saturating performance leading to a smaller gap, but also demonstrates that an image-based model may be good enough for existing video-language tasks.

\subsection{Video Question Answering}
\label{sec:vqa}

Compared with other CLIP-based methods which focus on alignment, our method provides a more general video foundation model by using video and text masking, which fuses features across modalities.
To validate this, we conduct experiments on video question answering. 
Unlike video-text retrieval which is purely similarity-based, question answering requires more interactions between modalities to predict the answer.

\noindent\textbf{Datasets.}
We report results on MSRVTT-QA~\cite{xu2017video}, MSVD-QA~\cite{xu2017video}, and the frame-QA subtask on TGIF-QA~\cite{tgif_qa}. MSRVTT-QA contains 243K open-ended questions over 10K videos. MSVD-QA consists of 47K open-ended questions over 2K videos. We follow the settings in All-in-one~\cite{all_in_one} as the dataset setup. Specifically, we choose 1,500, 1,000, and 1,540 most common options as the target vocabulary for each dataset respectively. 

\noindent\textbf{Implementation.}
We add a two-layer MLP on top of the pretrained model to predict the answer.
We consider three possible features to feed into the VQA classification head:
1) Alignment features only, which concatenates the classification tokens of the vision encoder and text encoder. This is the default setting when post-pretraining without text masking.
2) Fusion features only, which takes the end-of-text token in the text decoder features as the final classification feature. 
3) The combination of alignment and fusion features, which we find to work best in the experiments and is consistent with the intuition: alignment features and fusion features are complementary to each other.

\noindent\textbf{Training.}
For all three datasets, we train the post-pretrained model on standard training split with a learning rate of $1$$\times $$10^{-5}$ for 20 epochs. We use a cosine learning rate scheduler with 2 warm-up epochs. This simple choice of training hyperparameters shows that our post-pretraining framework produces a robust and general model.

We report the results of video question answering in Table~\ref{tab:vqa}. With the text masking, our method exhibits superior performance comparable to or even surpassing some methods designed specifically for VQA tasks such as Just Ask~\cite{yang2021just}. Compared with video foundation models counterparts, our method also performs better than some models trained on a large-scale dataset with heavy computational resources such as MERLOT~\cite{merlot} and All-in-one~\cite{all_in_one}. The text masking improves the accuracy by 0.6\%, 1.7\%, and 2.1\% on MSRVTT, MSVD, and TGIF-QA respectively, showing that the text masking design is effective in fusing different modalities.

We also attribute this performance gain to the better ability to model text. Currently, common video question answering practices share the same settings with classification, but with a much larger vocabulary size (e.g., 1,500 for MSRVTT-QA). Therefore, a well-trained text encoder is crucial for better performance. Similar observations are shared in Co-Tokenization~\cite{piergiovanni2022video}, FrozenBiLM~\cite{yang2022zero}, and Img2Prompt~\cite{guo2022images}, where frozen large language models are found to be extremely helpful in question answering tasks.

\begin{table}
    \small
    \centering
    \begin{tabular}[htop]{l c c c c}
        \toprule
        Method  & MSRVTT & MSVD & TGIF-QA \\
        \midrule
        \color{gray}{Just Ask~\cite{yang2021just}} & \color{gray}{41.8} & \color{gray}{47.5} & \color{gray}{-} \\
        \color{gray}{Co-Tokenization~\cite{piergiovanni2022video}} & \color{gray}{45.7} & \color{gray}{48.6} & \color{gray}{62.5} \\
        ClipBERT~\cite{clipbert} & 37.4 & - &  60.3 \\
        ALPRO~\cite{li2022align}  &  42.1 & 46.3 & - \\
        All-in-one~\cite{all_in_one} & 42.9 & 46.5 & 64.2 \\
        MERLOT~\cite{merlot} & 43.1 & - & \underline{69.5} \\
        VIOLET~\cite{violet} & 43.9 & 47.9 & 68.9 \\
        OmniVL~\cite{wang2022omnivl} & 44.1 & 51.0 & - \\
        VIOLETv2~\cite{fu2022empirical} & \underline{44.5} & \textbf{54.7} & \textbf{72.8} \\
        \hline
        Ours w/o decoder & 44.2 & 50.7 & 67.2 \\
        Ours with ViT & 44.1 & 50.1 & 67.0 \\
        Ours & \textbf{44.8} & \underline{52.4} & 69.3 \\
        \bottomrule
    \end{tabular}
    \caption{\textbf{Video question answering on MSRVTT, MSVD, and TGIF-QA.} Our method is aimed to compare with video foundation models for general purposes. Therefore those methods specially designed for video question answering are {\color{gray}{grayed out}}.
    }
    \vspace{-0.3cm}
    \label{tab:vqa}
\end{table}


\subsection{Ablation Study}
\label{sec:ablation}

\noindent\textbf{Drop Ratio of Video Patch Dropping.} 
We follow VideoMAE~\cite{tong2022videomae} with a drop ratio of $90\%$ when implementing video patch dropping. 
As retrieval task benefits most from patch dropping, we compare different dropping ratios with the same batch size in post-pretraining and show how the dropping ratio affects the downstream performance. 
The results on MSRVTT video-text retrieval are shown in Table~\ref{tab:abvtr}. 
As expected, when post-pretraining with the same batch size, a lower drop ratio yields higher performance due to the smaller gap between post-pretraining and downstream fine-tuning. 
However, the GPU memory usage of drop ratio 0.7 is 2.3 times higher than that of drop ratio 0.9. 
Intuitively there is a trade-off between the performance and efficiency of video patch dropping.
One should consider the efficiency and performance based on the limitation of computational resources.
We simply adopt the drop ratio of 0.9 for the highest efficiency with an acceptable performance drop on downstream tasks.
This also aligns with the conclusion of VideoMAE~\cite{tong2022videomae} that videos in nature are highly redundant. 
Videos can endure higher drop ratios compared to images which typically utilize a drop ratio around $70\%$~\cite{li2022scaling}.

\begin{table} 
    \centering
    \small
    \setlength\tabcolsep{3pt}
    \begin{tabular}[htop]{c c c c c c}
        \toprule
        Drop Ratio & R@1$\uparrow$ & R@5$\uparrow$ & R@10$\uparrow$ & MdR$\downarrow$  & Mem/G$\downarrow$ \\
        \midrule
        0.7 & \textbf{48.0} & \textbf{74.3} & \textbf{83.4} & \textbf{2.0} & 37.3 \\
        0.8 & 47.7 & 74.0 & 83.2 & \textbf{2.0} & 25.6\\
        \rowcolor{gray!20} 
        0.9 & 47.4 & 73.2 & 82.6 & \textbf{2.0} & \textbf{16.2}\\
        \bottomrule
    \end{tabular}
    \caption{\textbf{Different patch drop ratio.} Performance of our method on fine-tuned MSRVTT video-text retrieval when post-pretrained with different drop ratios for patch dropping. ``Mem" denotes single GPU memory usage with per GPU batch size 128.}
    \vspace{0.27cm}
    \label{tab:abvtr}
\end{table}

\begin{table}
    
    \centering
    \begin{tabular}[htop]{c c c c c}
        \toprule
        Mask Ratio & \color{gray}{0} & 0.05 & \cellcolor{gray!20}{0.15} & 0.25 \\
        \midrule
        Accuracy & \color{gray}{44.2} & 44.4 & \cellcolor{gray!20}{\textbf{44.8}} & 44.6 \\ 
        \bottomrule
    \end{tabular}
    \caption{\textbf{Differnt text masking ratio.} Performance of our method on MSRVTT video question answering when trained with different mask ratios for masked language module. ``0'' means removing the masked language module.}
    \vspace{-0.3cm}
    \label{tab:abmlm}
\end{table}
\noindent\textbf{Mask Ratio of Text Masking.} 
The default text masking ratio follows BERT~\cite{devlin2018bert} using $15\%$. 
However, considering that the masked text decoder serves as a cross-modal fuser, a different mask ratio may work better. 
As video question answering task benefits most from text masking, we compare different mask ratios in post-pretraining and evaluate on MSRVTT-QA. 
The results in Table~\ref{tab:abmlm} vary little on video question answering task. 
We attribute this to three reasons. 
First, a frozen text encoder fixes unmasked text features and leaves little space for the masked text decoder to learn. 
Second, the masked text decoder is not initialized with a pretrained model and is only post-pretrained for a relatively short schedule.
Third, the masked text decoder is jointly trained with patch dropping. With a high mask ratio, the decoder does not benefit from auxiliary vision information.



\begin{table}
    \centering
    \small
    \begin{tabular}[htop]{l r c c c}
        \toprule
        $k$ & $l$ & MSR-VTT & LSMDC & combined\\
        \midrule
        \color{gray}{-} & \color{gray}{-} & \color{gray}{92.6} & \color{gray}{74.9} & \color{gray}{78.9} \\
        2 & 50 & 93.1 & 75.0 & 79.1 \\
        2 & 10 & 93.4 & 76.0 & 80.0 \\
        \rowcolor{gray!20} 
        3 & 10 & 93.2 & \textbf{76.3} & \textbf{80.2} \\
        5 & 5 & \textbf{93.7} & 75.7 & 79.8 \\
        5 & 25 & 93.0 & 75.5 & 79.5\\
        \bottomrule
    \end{tabular}
    \caption{Performance of modified WiSE-FT on zero-shot multiple choice tasks with different training hyperparameters. 
    ``combined'' indicates the combination of two datasets. 
    When $k=2$ and $l=50$, this is standard WiSE-FT with $\alpha=0.5$.
    Our method is robust to the ensembling interval $k$ and the checkpoint number $l$.
    The modified version provides consistent improvement over the original WiSE-FT.
    }
    \vspace{-0.3cm}
    \label{tab:abwiserft}
\end{table}

\noindent\textbf{Hyperparameters of WiSE-FT.}
We use a modified version of WiSE-FT~\cite{wortsman2022robust} to further alleviate distribution shift in zero-shot tasks. 
To ablate this choice, we vary the ensembling interval $k$ and the number of checkpoints $l$ in post-pretraining.
As only zero-shot tasks benefit most from WiSE-FT, we conduct ablation analysis on $k$ and $l$ on zero-shot multiple-choice tasks.
The results are shown in Table~\ref{tab:abwiserft}.
The results reveal that the zero-shot performance does not vary drastically with different training hyperparameters, but the modified version provides a consistent improvement over the original WiSE-FT.

\subsection{Discussion}
\label{sec:discussion}

\noindent\textbf{Reflections on Video-Language Training.}
Our results provide several observations on current video-language training, including pretraining data and downstream tasks. In short, they may not be ``video'' enough. 
\textbf{(i)} Our method is a simple post-pretraining framework without bells and whistles and is trained with minimal costs. 
Still, it achieves performance comparable to heavily trained video foundation models. We give credit to the powerful CLIP pretraining. 
Several studies have already found that image-based models can perform well on video benchmarks including CoCa~\cite{yu2022coca} and Singularity~\cite{singularity}. 
Also in CLIP-ViP~\cite{xue2022clip}, video-language pretraining can be improved with captions generated by an image captioner.
\textbf{(ii)} Our experiments are conducted with two different types of backbones: UniformerV2 and vanilla ViT. 
One would expect that a powerful spatiotemporal backbone like UniformerV2 will outperform vanilla ViT to a large extent, but in our study this is not the case. 
The vanilla ViT even surpasses UniformerV2 on several downstream tasks, mostly in zero-shot settings. 
The performance gap between ViT and UniformerV2 on retrieval tasks also shrinks from zero-shot to finetuning setting.
This indicates that temporal modeling may not be so important on some of current video-language benchmarks. 
Or spatiotemporal backbones like UniformerV2 have not fully utilized temporal information yet.
\textbf{(iii)} Videos should intuitively contain more information than images, but commonly used video-text data like WebVid does not demonstrate longer and richer text descriptions compared with image-text data. This is also one of the reasons why a frozen text encoder is good enough in our method.

\noindent\textbf{The Role of Text Encoder.}
Our method will perform worse if the text encoder is not frozen, which shares the same observation with CLIP-ViP.
CLIP-ViP attributes this to language domain gap and tackles this problem by using extensively generated caption data, while our method simply freezes the text encoder.
We believe that a well-trained text encoder can further boost the performance on video-language tasks.
This has also been demonstrated by some works on question answering with the help of powerful language models including Co-Tokenization~\cite{piergiovanni2022video}, FrozenBiLM~\cite{yang2022zero}, and Img2Prompt~\cite{guo2022images}.
However, current video-text datasets may not be diverse and high-quality enough to train a satisfying text encoder, especially when adopting a well-pretrained model like CLIP. 
Further post-pretraining without sufficient text data will hurt the performance.

\noindent\textbf{Future Directions}
We provide two possible directions of improvement based on our observations. 
The first one is to build models and benchmarks that are more ``video'' with more temporal reasoning. 
For example, a good video model should be able to distinguish between ``a plane is taking off'' and ``a plane is landing'' (which unfortunately, most models cannot), and a good video benchmark should focus more on temporal information. 
The second one is to form richer language descriptions for video-text data. 
For example, dense captions containing much more textual information with explicit timestamps may be more beneficial to temporal modeling than the current ``one caption per video'' setting.

\section{Conclusion}
\label{sec:conclusion}

We propose a simple yet efficient post-pretraining framework to build video foundation models based on image foundation models.
With the introduction of video patch dropping and text masking, our method achieves state-of-the-art performance on various video-language tasks, including various zero-shot tasks, video question answering, and video-text retrieval.
The performances of our model are superior to some heavily pretrained video foundation models.
The experimental results demonstrate the effectiveness and generality of our method.
Our method establishes a novel counterpart for video foundation models and provides in-depth reflections on video-language training and the role of the text encoder.
We hope our method can provide a new direction in building large pretrained models, making them more accessible and sustainable.

\noindent\textbf{Societal Impacts.} Due to the efficiency of our method, it enables large pretrained models to be more accessible for small research organizations and more environmentally friendly with less carbon footprint, which is one of the major concerns of large pretrained models. However, our method shares the same potential negative impact as the CLIP model does, a zero-shot classification can be used for surveillance especially now it is the engifted ability to classify temporal actions. Also, a post-pretraining procedure may make it hard to trace data sources when handling copyright or privacy issues.

{\small
\bibliographystyle{ieee_fullname}
\bibliography{egbib}

@String(CVPR= {IEEE Conf. Comput. Vis. Pattern Recog.})

@String(ICCV= {Int. Conf. Comput. Vis.})

@String(ECCV= {Eur. Conf. Comput. Vis.})

@String(ICLR = {Int. Conf. Learn. Represent.})

@String(AAAI = {AAAI})

@String(CVPR  = {CVPR})

@String(ICCV  = {ICCV})

@String(ECCV  = {ECCV})

@String(ICLR  = {ICLR})

@inproceedings{radford2021learning,
  title={Learning transferable visual models from natural language supervision},
  author={Radford, Alec and Kim, Jong Wook and Hallacy, Chris and Ramesh, Aditya and Goh, Gabriel and Agarwal, Sandhini and Sastry, Girish and Askell, Amanda and Mishkin, Pamela and Clark, Jack and others},
  booktitle={International Conference on Machine Learning},
  pages={8748--8763},
  year={2021},
  organization={PMLR}
}

@article{violet,
  title={VIOLET: End-to-end video-language transformers with masked visual-token modeling},
  author={Fu, Tsu-Jui and Li, Linjie and Gan, Zhe and Lin, Kevin and Wang, William Yang and Wang, Lijuan and Liu, Zicheng},
  journal={arXiv preprint arXiv:2111.12681},
  year={2021}
}

@article{all_in_one,
  title={All in one: Exploring unified video-language pre-training},
  author={Wang, Alex Jinpeng and Ge, Yixiao and Yan, Rui and Ge, Yuying and Lin, Xudong and Cai, Guanyu and Wu, Jianping and Shan, Ying and Qie, Xiaohu and Shou, Mike Zheng},
  journal={arXiv preprint arXiv:2203.07303},
  year={2022}
}

@article{lavender,
  title={Lavender: Unifying video-language understanding as masked language modeling},
  author={Li, Linjie and Gan, Zhe and Lin, Kevin and Lin, Chung-Ching and Liu, Zicheng and Liu, Ce and Wang, Lijuan},
  journal={arXiv preprint arXiv:2206.07160},
  year={2022}
}

@inproceedings{reserve,
  title={Merlot reserve: Neural script knowledge through vision and language and sound},
  author={Zellers, Rowan and Lu, Jiasen and Lu, Ximing and Yu, Youngjae and Zhao, Yanpeng and Salehi, Mohammadreza and Kusupati, Aditya and Hessel, Jack and Farhadi, Ali and Choi, Yejin},
  booktitle={Proceedings of the IEEE/CVF Conference on Computer Vision and Pattern Recognition},
  pages={16375--16387},
  year={2022}
}

@article{merlot,
  title={Merlot: Multimodal neural script knowledge models},
  author={Zellers, Rowan and Lu, Ximing and Hessel, Jack and Yu, Youngjae and Park, Jae Sung and Cao, Jize and Farhadi, Ali and Choi, Yejin},
  journal={Advances in Neural Information Processing Systems},
  volume={34},
  pages={23634--23651},
  year={2021}
}

@article{videoclip,
  title={Videoclip: Contrastive pre-training for zero-shot video-text understanding},
  author={Xu, Hu and Ghosh, Gargi and Huang, Po-Yao and Okhonko, Dmytro and Aghajanyan, Armen and Metze, Florian and Zettlemoyer, Luke and Feichtenhofer, Christoph},
  journal={arXiv preprint arXiv:2109.14084},
  year={2021}
}

@inproceedings{clipbert,
  title={Less is more: Clipbert for video-and-language learning via sparse sampling},
  author={Lei, Jie and Li, Linjie and Zhou, Luowei and Gan, Zhe and Berg, Tamara L and Bansal, Mohit and Liu, Jingjing},
  booktitle={Proceedings of the IEEE/CVF Conference on Computer Vision and Pattern Recognition},
  pages={7331--7341},
  year={2021}
}

@article{videobert,
  title={VideoBERT: A Joint Model for Video and Language Representation Learning},
  author={Chen Sun and Austin Myers and Carl Vondrick and Kevin P. Murphy and Cordelia Schmid},
  journal={2019 IEEE/CVF International Conference on Computer Vision (ICCV)},
  year={2019},
  pages={7463-7472}
}

@article{actbert,
  title={ActBERT: Learning Global-Local Video-Text Representations},
  author={Linchao Zhu and Yi Yang},
  journal={2020 IEEE/CVF Conference on Computer Vision and Pattern Recognition (CVPR)},
  year={2020},
  pages={8743-8752}
}

@inproceedings{frozen,
  title={Frozen in time: A joint video and image encoder for end-to-end retrieval},
  author={Bain, Max and Nagrani, Arsha and Varol, G{\"u}l and Zisserman, Andrew},
  booktitle={Proceedings of the IEEE/CVF International Conference on Computer Vision},
  pages={1728--1738},
  year={2021}
}

@article{singularity,
  title={Revealing Single Frame Bias for Video-and-Language Learning},
  author={Jie Lei and Tamara L. Berg and Mohit Bansal},
  journal={ArXiv},
  year={2022},
  volume={abs/2206.03428}
}

@inproceedings{uniter,
  title={UNITER: UNiversal Image-TExt Representation Learning},
  author={Yen-Chun Chen and Linjie Li and Licheng Yu and Ahmed El Kholy and Faisal Ahmed and Zhe Gan and Yu Cheng and Jingjing Liu},
  booktitle={ECCV},
  year={2020}
}

@inproceedings{LXMERT,
  title={LXMERT: Learning Cross-Modality Encoder Representations from Transformers},
  author={Hao Hao Tan and Mohit Bansal},
  booktitle={EMNLP},
  year={2019},
}

@inproceedings{vl_bert,
  title={VL-BERT: Pre-training of Generic Visual-Linguistic Representations},
  author={Weijie Su and Xizhou Zhu and Yue Cao and Bin Li and Lewei Lu and Furu Wei and Jifeng Dai},
  booktitle={ICLR},
  year={2020},
}

@article{value,
  title={Value: A multi-task benchmark for video-and-language understanding evaluation},
  author={Li, Linjie and Lei, Jie and Gan, Zhe and Yu, Licheng and Chen, Yen-Chun and Pillai, Rohit and Cheng, Yu and Zhou, Luowei and Wang, Xin Eric and Wang, William Yang and others},
  journal={arXiv preprint arXiv:2106.04632},
  year={2021}
}

@article{liu2019use,
  title={Use what you have: Video retrieval using representations from collaborative experts},
  author={Liu, Yang and Albanie, Samuel and Nagrani, Arsha and Zisserman, Andrew},
  journal={arXiv preprint arXiv:1907.13487},
  year={2019}
}

@inproceedings{jiang2020divide,
  title={Divide and conquer: Question-guided spatio-temporal contextual attention for video question answering},
  author={Jiang, Jianwen and Chen, Ziqiang and Lin, Haojie and Zhao, Xibin and Gao, Yue},
  booktitle={Proceedings of the AAAI Conference on Artificial Intelligence},
  volume={34},
  number={07},
  pages={11101--11108},
  year={2020}
}

@inproceedings{le2020hierarchical,
  title={Hierarchical conditional relation networks for video question answering},
  author={Le, Thao Minh and Le, Vuong and Venkatesh, Svetha and Tran, Truyen},
  booktitle={Proceedings of the IEEE/CVF conference on computer vision and pattern recognition},
  pages={9972--9981},
  year={2020}
}

@article{patrick2020support,
  title={Support-set bottlenecks for video-text representation learning},
  author={Patrick, Mandela and Huang, Po-Yao and Asano, Yuki and Metze, Florian and Hauptmann, Alexander and Henriques, Joao and Vedaldi, Andrea},
  journal={arXiv preprint arXiv:2010.02824},
  year={2020}
}

@inproceedings{gabeur2020multi,
  title={Multi-modal transformer for video retrieval},
  author={Gabeur, Valentin and Sun, Chen and Alahari, Karteek and Schmid, Cordelia},
  booktitle={European Conference on Computer Vision},
  pages={214--229},
  year={2020},
  organization={Springer}
}

@inproceedings{tgif_qa,
  title={Tgif-qa: Toward spatio-temporal reasoning in visual question answering},
  author={Jang, Yunseok and Song, Yale and Yu, Youngjae and Kim, Youngjin and Kim, Gunhee},
  booktitle={Proceedings of the IEEE conference on computer vision and pattern recognition},
  pages={2758--2766},
  year={2017}
}

@article{tvqa,
  title={Tvqa: Localized, compositional video question answering},
  author={Lei, Jie and Yu, Licheng and Bansal, Mohit and Berg, Tamara L},
  journal={arXiv preprint arXiv:1809.01696},
  year={2018}
}

@article{tvqa+,
  title={Tvqa+: Spatio-temporal grounding for video question answering},
  author={Lei, Jie and Yu, Licheng and Berg, Tamara L and Bansal, Mohit},
  journal={arXiv preprint arXiv:1904.11574},
  year={2019}
}

@inproceedings{xu2017video,
  title={Video question answering via gradually refined attention over appearance and motion},
  author={Xu, Dejing and Zhao, Zhou and Xiao, Jun and Wu, Fei and Zhang, Hanwang and He, Xiangnan and Zhuang, Yueting},
  booktitle={Proceedings of the 25th ACM international conference on Multimedia},
  pages={1645--1653},
  year={2017}
}

@inproceedings{msrvtt,
  title={Msr-vtt: A large video description dataset for bridging video and language},
  author={Xu, Jun and Mei, Tao and Yao, Ting and Rui, Yong},
  booktitle={Proceedings of the IEEE conference on computer vision and pattern recognition},
  pages={5288--5296},
  year={2016}
}

@inproceedings{densecaption,
  title={Dense-captioning events in videos},
  author={Krishna, Ranjay and Hata, Kenji and Ren, Frederic and Fei-Fei, Li and Carlos Niebles, Juan},
  booktitle={Proceedings of the IEEE international conference on computer vision},
  pages={706--715},
  year={2017}
}

@inproceedings{tvr,
  title={Tvr: A large-scale dataset for video-subtitle moment retrieval},
  author={Lei, Jie and Yu, Licheng and Berg, Tamara L and Bansal, Mohit},
  booktitle={European Conference on Computer Vision},
  pages={447--463},
  year={2020},
  organization={Springer}
}

@inproceedings{vatex,
  title={Vatex: A large-scale, high-quality multilingual dataset for video-and-language research},
  author={Wang, Xin and Wu, Jiawei and Chen, Junkun and Li, Lei and Wang, Yuan-Fang and Wang, William Yang},
  booktitle={Proceedings of the IEEE/CVF International Conference on Computer Vision},
  pages={4581--4591},
  year={2019}
}

@inproceedings{zhou2018towards,
  title={Towards automatic learning of procedures from web instructional videos},
  author={Zhou, Luowei and Xu, Chenliang and Corso, Jason J},
  booktitle={Thirty-Second AAAI Conference on Artificial Intelligence},
  year={2018}
}

@inproceedings{movie_description,
  title={A dataset for movie description},
  author={Rohrbach, Anna and Rohrbach, Marcus and Tandon, Niket and Schiele, Bernt},
  booktitle={Proceedings of the IEEE conference on computer vision and pattern recognition},
  pages={3202--3212},
  year={2015}
}

@inproceedings{didemo,
  title={Localizing moments in video with natural language},
  author={Anne Hendricks, Lisa and Wang, Oliver and Shechtman, Eli and Sivic, Josef and Darrell, Trevor and Russell, Bryan},
  booktitle={Proceedings of the IEEE international conference on computer vision},
  pages={5803--5812},
  year={2017}
}

@inproceedings{Tall,
  title={Tall: Temporal activity localization via language query},
  author={Gao, Jiyang and Sun, Chen and Yang, Zhenheng and Nevatia, Ram},
  booktitle={Proceedings of the IEEE international conference on computer vision},
  pages={5267--5275},
  year={2017}
}

@inproceedings{xie2018rethinking,
  title={Rethinking spatiotemporal feature learning: Speed-accuracy trade-offs in video classification},
  author={Xie, Saining and Sun, Chen and Huang, Jonathan and Tu, Zhuowen and Murphy, Kevin},
  booktitle={Proceedings of the European conference on computer vision (ECCV)},
  pages={305--321},
  year={2018}
}

@inproceedings{gao2018motion,
  title={Motion-appearance co-memory networks for video question answering},
  author={Gao, Jiyang and Ge, Runzhou and Chen, Kan and Nevatia, Ram},
  booktitle={Proceedings of the IEEE Conference on Computer Vision and Pattern Recognition},
  pages={6576--6585},
  year={2018}
}

@article{lei2021detecting,
  title={Detecting Moments and Highlights in Videos via Natural Language Queries},
  author={Lei, Jie and Berg, Tamara L and Bansal, Mohit},
  journal={Advances in Neural Information Processing Systems},
  volume={34},
  pages={11846--11858},
  year={2021}
}

@inproceedings{fan2019heterogeneous,
  title={Heterogeneous memory enhanced multimodal attention model for video question answering},
  author={Fan, Chenyou and Zhang, Xiaofan and Zhang, Shu and Wang, Wensheng and Zhang, Chi and Huang, Heng},
  booktitle={Proceedings of the IEEE/CVF conference on computer vision and pattern recognition},
  pages={1999--2007},
  year={2019}
}

@inproceedings{msvd,
  title={Collecting highly parallel data for paraphrase evaluation},
  author={Chen, David and Dolan, William B},
  booktitle={Proceedings of the 49th annual meeting of the association for computational linguistics: human language technologies},
  pages={190--200},
  year={2011}
}

@inproceedings{CC3M,
  title={Conceptual Captions: A Cleaned, Hypernymed, Image Alt-text Dataset For Automatic Image Captioning},
  author={Piyush Sharma and Nan Ding and Sebastian Goodman and Radu Soricut},
  booktitle={ACL},
  year={2018}
}

@article{coco,
  title={Microsoft COCO Captions: Data Collection and Evaluation Server},
  author={Xinlei Chen and Hao Fang and Tsung-Yi Lin and Ramakrishna Vedantam and Saurabh Gupta and Piotr Doll{\'a}r and C. Lawrence Zitnick},
  journal={ArXiv},
  year={2015},
  volume={abs/1504.00325}
}

@article{vg,
  title={Visual Genome: Connecting Language and Vision Using Crowdsourced Dense Image Annotations},
  author={Ranjay Krishna and Yuke Zhu and Oliver Groth and Justin Johnson and Kenji Hata and Joshua Kravitz and Stephanie Chen and Yannis Kalantidis and Li-Jia Li and David A. Shamma and Michael S. Bernstein and Li Fei-Fei},
  journal={International Journal of Computer Vision},
  year={2016},
  volume={123},
  pages={32-73}
}

@article{howto100m,
  title={HowTo100M: Learning a Text-Video Embedding by Watching Hundred Million Narrated Video Clips},
  author={Antoine Miech and Dimitri Zhukov and Jean-Baptiste Alayrac and Makarand Tapaswi and Ivan Laptev and Josef Sivic},
  journal={2019 IEEE/CVF International Conference on Computer Vision (ICCV)},
  year={2019},
  pages={2630-2640}
}

@article{hero,
  title={Hero: Hierarchical encoder for video+ language omni-representation pre-training},
  author={Li, Linjie and Chen, Yen-Chun and Cheng, Yu and Gan, Zhe and Yu, Licheng and Liu, Jingjing},
  journal={arXiv preprint arXiv:2005.00200},
  year={2020}
}

@article{tip-adapter,
  title={Tip-adapter: Training-free clip-adapter for better vision-language modeling},
  author={Zhang, Renrui and Fang, Rongyao and Gao, Peng and Zhang, Wei and Li, Kunchang and Dai, Jifeng and Qiao, Yu and Li, Hongsheng},
  journal={arXiv preprint arXiv:2111.03930},
  year={2021}
}

@article{clip4clip,
  title={CLIP4Clip: An empirical study of CLIP for end to end video clip retrieval and captioning},
  author={Luo, Huaishao and Ji, Lei and Zhong, Ming and Chen, Yang and Lei, Wen and Duan, Nan and Li, Tianrui},
  journal={Neurocomputing},
  volume={508},
  pages={293--304},
  year={2022},
  publisher={Elsevier}
}

@inproceedings{wortsman2022robust,
  title={Robust fine-tuning of zero-shot models},
  author={Wortsman, Mitchell and Ilharco, Gabriel and Kim, Jong Wook and Li, Mike and Kornblith, Simon and Roelofs, Rebecca and Lopes, Raphael Gontijo and Hajishirzi, Hannaneh and Farhadi, Ali and Namkoong, Hongseok and others},
  booktitle={Proceedings of the IEEE/CVF Conference on Computer Vision and Pattern Recognition},
  pages={7959--7971},
  year={2022}
}

@article{xue2022clip,
  title={CLIP-ViP: Adapting Pre-trained Image-Text Model to Video-Language Representation Alignment},
  author={Xue, Hongwei and Sun, Yuchong and Liu, Bei and Fu, Jianlong and Song, Ruihua and Li, Houqiang and Luo, Jiebo},
  journal={arXiv preprint arXiv:2209.06430},
  year={2022}
}

@inproceedings{xue2022advancing,
  title={Advancing High-Resolution Video-Language Representation with Large-Scale Video Transcriptions},
  author={Xue, Hongwei and Hang, Tiankai and Zeng, Yanhong and Sun, Yuchong and Liu, Bei and Yang, Huan and Fu, Jianlong and Guo, Baining},
  booktitle={Proceedings of the IEEE/CVF Conference on Computer Vision and Pattern Recognition},
  pages={5036--5045},
  year={2022}
}

@article{li2021align,
  title={Align before fuse: Vision and language representation learning with momentum distillation},
  author={Li, Junnan and Selvaraju, Ramprasaath and Gotmare, Akhilesh and Joty, Shafiq and Xiong, Caiming and Hoi, Steven Chu Hong},
  journal={Advances in neural information processing systems},
  volume={34},
  pages={9694--9705},
  year={2021}
}

@article{yu2022coca,
  title={Coca: Contrastive captioners are image-text foundation models},
  author={Yu, Jiahui and Wang, Zirui and Vasudevan, Vijay and Yeung, Legg and Seyedhosseini, Mojtaba and Wu, Yonghui},
  journal={arXiv preprint arXiv:2205.01917},
  year={2022}
}

@article{tong2022videomae,
  title={Videomae: Masked autoencoders are data-efficient learners for self-supervised video pre-training},
  author={Tong, Zhan and Song, Yibing and Wang, Jue and Wang, Limin},
  journal={arXiv preprint arXiv:2203.12602},
  year={2022}
}

@inproceedings{he2022masked,
  title={Masked autoencoders are scalable vision learners},
  author={He, Kaiming and Chen, Xinlei and Xie, Saining and Li, Yanghao and Doll{\'a}r, Piotr and Girshick, Ross},
  booktitle={Proceedings of the IEEE/CVF Conference on Computer Vision and Pattern Recognition},
  pages={16000--16009},
  year={2022}
}

@article{han2022turbo,
  title={Turbo Training with Token Dropout},
  author={Han, Tengda and Xie, Weidi and Zisserman, Andrew},
  journal={arXiv preprint arXiv:2210.04889},
  year={2022}
}

@article{anonymous2023uniformerv,
  title={UniFormerV2: Spatiotemporal Learning by Arming Image ViTs with Video UniFormer},
  author={Li, Kunchang and Wang, Yali and He, Yinan and Li, Yizhuo and Wang, Yi and Wang, Limin and Qiao, Yu},
  journal={arXiv preprint arXiv:2211.09552},
  year={2022}
}

@inproceedings{jia2021scaling,
  title={Scaling up visual and vision-language representation learning with noisy text supervision},
  author={Jia, Chao and Yang, Yinfei and Xia, Ye and Chen, Yi-Ting and Parekh, Zarana and Pham, Hieu and Le, Quoc and Sung, Yun-Hsuan and Li, Zhen and Duerig, Tom},
  booktitle={International Conference on Machine Learning},
  pages={4904--4916},
  year={2021},
  organization={PMLR}
}

@article{coop,
  title={Learning to Prompt for Vision-Language Models},
  author={Kaiyang Zhou and Jingkang Yang and Chen Change Loy and Ziwei Liu},
  journal={Int. J. Comput. Vis.},
  year={2022},
  volume={130},
  pages={2337-2348}
}

@article{cocoop,
  title={Conditional Prompt Learning for Vision-Language Models},
  author={Kaiyang Zhou and Jingkang Yang and Chen Change Loy and Ziwei Liu},
  journal={2022 IEEE/CVF Conference on Computer Vision and Pattern Recognition (CVPR)},
  year={2022},
  pages={16795-16804}
}

@article{clip2tv,
  title={CLIP2TV: An Empirical Study on Transformer-based Methods for Video-Text Retrieval},
  author={Zijian Gao and Jingyun Liu and Sheng Chen and Dedan Chang and Hao Zhang and Jinwei Yuan},
  journal={ArXiv},
  year={2021},
  volume={abs/2111.05610}
}

@article{clip2video,
  title={CLIP2Video: Mastering Video-Text Retrieval via Image CLIP},
  author={Han Fang and Pengfei Xiong and Luhui Xu and Yu Chen},
  journal={ArXiv},
  year={2021},
  volume={abs/2106.11097}
}

@article{camoe,
  title={Improving Video-Text Retrieval by Multi-Stream Corpus Alignment and Dual Softmax Loss},
  author={Xingyi Cheng and Hezheng Lin and Xiangyu Wu and F. Yang and Dong Shen},
  journal={ArXiv},
  year={2021},
  volume={abs/2109.04290}
}

@article{clip4caption,
  title={CLIP4Caption: CLIP for Video Caption},
  author={Mingkang Tang and Zhanyu Wang and Zhenhua Liu and Fengyun Rao and Dian Li and Xiu Li},
  journal={Proceedings of the 29th ACM International Conference on Multimedia},
  year={2021}
}

@inproceedings{clipit,
  title={CLIP-It! Language-Guided Video Summarization},
  author={Medhini Narasimhan and Anna Rohrbach and Trevor Darrell},
  booktitle={NeurIPS},
  year={2021}
}

@article{clip_meet_video,
  title={CLIP Meets Video Captioners: Attribute-Aware Representation Learning Promotes Accurate Captioning},
  author={Bang Yang and Yuexian Zou},
  journal={ArXiv},
  year={2021},
  volume={abs/2111.15162}
}

@article{actionclip,
  title={ActionCLIP: A New Paradigm for Video Action Recognition},
  author={Mengmeng Wang and Jiazheng Xing and Yong Liu},
  journal={ArXiv},
  year={2021},
  volume={abs/2109.08472}
}

@inproceedings{chen2021elaborative,
  title={Elaborative rehearsal for zero-shot action recognition},
  author={Chen, Shizhe and Huang, Dong},
  booktitle={Proceedings of the IEEE/CVF International Conference on Computer Vision},
  pages={13638--13647},
  year={2021}
}

@inproceedings{yu2018joint,
  title={A joint sequence fusion model for video question answering and retrieval},
  author={Yu, Youngjae and Kim, Jongseok and Kim, Gunhee},
  booktitle={Proceedings of the European Conference on Computer Vision (ECCV)},
  pages={471--487},
  year={2018}
}

@inproceedings{li2022align,
  title={Align and Prompt: Video-and-Language Pre-training with Entity Prompts},
  author={Li, Dongxu and Li, Junnan and Li, Hongdong and Niebles, Juan Carlos and Hoi, Steven CH},
  booktitle={Proceedings of the IEEE/CVF Conference on Computer Vision and Pattern Recognition},
  pages={4953--4963},
  year={2022}
}

@inproceedings{ge2022bridging,
  title={Bridging Video-Text Retrieval With Multiple Choice Questions},
  author={Ge, Yuying and Ge, Yixiao and Liu, Xihui and Li, Dian and Shan, Ying and Qie, Xiaohu and Luo, Ping},
  booktitle={Proceedings of the IEEE/CVF Conference on Computer Vision and Pattern Recognition},
  pages={16167--16176},
  year={2022}
}

@article{wang2022omnivl,
  title={Omnivl: One foundation model for image-language and video-language tasks},
  author={Wang, Junke and Chen, Dongdong and Wu, Zuxuan and Luo, Chong and Zhou, Luowei and Zhao, Yucheng and Xie, Yujia and Liu, Ce and Jiang, Yu-Gang and Yuan, Lu},
  journal={arXiv preprint arXiv:2209.07526},
  year={2022}
}

@inproceedings{amrani2021noise,
  title={Noise estimation using density estimation for self-supervised multimodal learning},
  author={Amrani, Elad and Ben-Ari, Rami and Rotman, Daniel and Bronstein, Alex},
  booktitle={Proceedings of the AAAI Conference on Artificial Intelligence},
  volume={35},
  number={8},
  pages={6644--6652},
  year={2021}
}

@inproceedings{yang2021just,
  title={Just ask: Learning to answer questions from millions of narrated videos},
  author={Yang, Antoine and Miech, Antoine and Sivic, Josef and Laptev, Ivan and Schmid, Cordelia},
  booktitle={Proceedings of the IEEE/CVF International Conference on Computer Vision},
  pages={1686--1697},
  year={2021}
}

@inproceedings{piergiovanni2022video,
  title={Video question answering with iterative video-text co-tokenization},
  author={Piergiovanni, AJ and Morton, Kairo and Kuo, Weicheng and Ryoo, Michael S and Angelova, Anelia},
  booktitle={European Conference on Computer Vision},
  pages={76--94},
  year={2022},
  organization={Springer}
}

@article{Smaira2020ASN,
  title={A Short Note on the Kinetics-700-2020 Human Action Dataset},
  author={Lucas Smaira and Jo{\~a}o Carreira and Eric Noland and Ellen Clancy and Amy Wu and Andrew Zisserman},
  journal={ArXiv},
  year={2020},
  volume={abs/2010.10864}
}

@article{lsmdc_mc,
  title={Learning Language-Visual Embedding for Movie Understanding with Natural-Language},
  author={Atousa Torabi and Niket Tandon and Leonid Sigal},
  journal={ArXiv},
  year={2016},
  volume={abs/1609.08124}
}

@article{devlin2018bert,
  title={Bert: Pre-training of deep bidirectional transformers for language understanding},
  author={Devlin, Jacob and Chang, Ming-Wei and Lee, Kenton and Toutanova, Kristina},
  journal={arXiv preprint arXiv:1810.04805},
  year={2018}
}

@article{transformer,
  title={Attention is all you need},
  author={Vaswani, Ashish and Shazeer, Noam and Parmar, Niki and Uszkoreit, Jakob and Jones, Llion and Gomez, Aidan N and Kaiser, {\L}ukasz and Polosukhin, Illia},
  journal={Advances in neural information processing systems},
  volume={30},
  year={2017}
}

@article{Dosovitskiy2021AnII,
  title={An Image is Worth 16x16 Words: Transformers for Image Recognition at Scale},
  author={Alexey Dosovitskiy and Lucas Beyer and Alexander Kolesnikov and Dirk Weissenborn and Xiaohua Zhai and Thomas Unterthiner and Mostafa Dehghani and Matthias Minderer and Georg Heigold and Sylvain Gelly and Jakob Uszkoreit and Neil Houlsby},
  journal={ArXiv},
  year={2021},
  volume={abs/2010.11929}
}

@article{li2022scaling,
  title={Scaling Language-Image Pre-training via Masking},
  author={Li, Yanghao and Fan, Haoqi and Hu, Ronghang and Feichtenhofer, Christoph and He, Kaiming},
  journal={arXiv preprint arXiv:2212.00794},
  year={2022}
}

@article{guo2022images,
  title={From Images to Textual Prompts: Zero-shot VQA with Frozen Large Language Models},
  author={Guo, Jiaxian and Li, Junnan and Li, Dongxu and Tiong, Anthony Meng Huat and Li, Boyang and Tao, Dacheng and Hoi, Steven CH},
  journal={arXiv preprint arXiv:2212.10846},
  year={2022}
}

@article{fu2022empirical,
  title={An empirical study of end-to-end video-language transformers with masked visual modeling},
  author={Fu, Tsu-Jui and Li, Linjie and Gan, Zhe and Lin, Kevin and Wang, William Yang and Wang, Lijuan and Liu, Zicheng},
  journal={arXiv preprint arXiv:2209.01540},
  year={2022}
}

@article{schuhmann2022laion,
  title={Laion-5b: An open large-scale dataset for training next generation image-text models},
  author={Schuhmann, Christoph and Beaumont, Romain and Vencu, Richard and Gordon, Cade and Wightman, Ross and Cherti, Mehdi and Coombes, Theo and Katta, Aarush and Mullis, Clayton and Wortsman, Mitchell and others},
  journal={arXiv preprint arXiv:2210.08402},
  year={2022}
}

@article{feichtenhofer2022masked,
  title={Masked autoencoders as spatiotemporal learners},
  author={Feichtenhofer, Christoph and Fan, Haoqi and Li, Yanghao and He, Kaiming},
  journal={arXiv preprint arXiv:2205.09113},
  year={2022}
}

@article{yang2022zero,
  title={Zero-shot video question answering via frozen bidirectional language models},
  author={Yang, Antoine and Miech, Antoine and Sivic, Josef and Laptev, Ivan and Schmid, Cordelia},
  journal={arXiv preprint arXiv:2206.08155},
  year={2022}
}
}

\end{document}